\documentclass[10pt,twocolumn,letterpaper]{article}

\usepackage[pagenumbers]{cvpr} 

\makeatletter
\@namedef{ver@everyshi.sty}{}
\makeatother

\usepackage{graphicx}
\usepackage{amsmath}
\usepackage{amssymb}
\usepackage{booktabs}
\usepackage{subfiles} 
\usepackage{tikz}
\usepackage{multirow}
\DeclareMathOperator*{\argmin}{arg\,min}
\definecolor{yd}{rgb}{0.0, 0.5, 0.0}

\usepackage[pagebackref,breaklinks,colorlinks]{hyperref}
\newcommand{\ydnote}[1]{\textcolor{red}{[{\bf YD:} #1]}}

\newcommand{\rdnote}[1]{\textcolor{blue}{[{\bf RD:} #1]}}

\newcommand{\onlykey}[1]{\textcolor{black}{{\textit{only-key}}}}
\newcommand{\noatt}[1]{\textcolor{black}{{\textit{no-attn}}}}
\newcommand{\expres}[1]{\textcolor{black}{{EXPRES}}}
\newcommand{\vtab}[1]{\textcolor{black}{{VTAB-1k}}}
\newcommand{\vpt}[1]{\textcolor{black}{{VPT}}}
\newcommand{\vptsh}[1]{\textcolor{black}{{VPT-shallow}}}

\newcommand{\partialft}[0]{\textsc{Partial}}
\newcommand{\linear}[0]{\textsc{Linear}}
\newcommand{\fullft}[0]{\textsc{Full}}
\newcommand{\sidetune}[0]{\textsc{Sidetune}}
\newcommand{\mlp}[0]{\textsc{Mlp}}
\newcommand{\bias}[0]{\textsc{Bias}}

\newcommand{\pascal}[1]{\textcolor{black}{{$\mathbf{\text{PASCAL}-5^#1}$}}}

\usepackage[capitalize]{cleveref}
\crefname{section}{Sec.}{Secs.}
\Crefname{section}{Section}{Sections}
\Crefname{table}{Table}{Tables}
\crefname{table}{Tab.}{Tabs.}

\def\cvprPaperID{12148} 
\def\confName{CVPR}
\def\confYear{2023}

\begin{document}

\title{Learning Expressive Prompting With Residuals for Vision Transformers}


\author{Rajshekhar Das$^{1,2*}$, Yonatan Dukler$^{2}$, Avinash Ravichandran$^{2**}$, Ashwin Swaminathan$^{2}$ \\
Carnegie Mellon University$^{1}$ \hspace{1cm} AWS AI Labs$^{2}$ \\
{\tt\small rajshekd@andrew.cmu.edu,~dukler@amazon.com,~swashwin@amazon.com}
}
\maketitle

\begin{abstract}
Prompt learning is an  efficient approach to adapt transformers by inserting learnable set of parameters into the input and intermediate representations of a pre-trained model.
In this work, we present Expressive Prompts with Residuals (\expres{}) which modifies the prompt learning paradigm specifically for effective adaptation of vision transformers (ViT).
Our method constructs downstream representations via learnable ``output'' tokens (\textit{shallow} prompts), that are akin to the learned class tokens of the ViT. 
Further for better steering of the downstream representation processed by the frozen transformer, we introduce residual learnable tokens that are added to the output of various computations. 
We apply \expres{} for image classification and few-shot semantic segmentation, and show our method is capable of achieving state of the art prompt tuning on 3/3 categories of the VTAB benchmark.
In addition to strong performance, we observe that our approach is an order of magnitude more prompt efficient than existing visual prompting baselines.
We analytically show the computational benefits of our approach over weight space adaptation  techniques like finetuning.
Lastly we systematically corroborate the architectural design of our method via a series of ablation experiments.

\let\thefootnote\relax\footnotetext{$^*$Work conducted while interning at AWS AI Labs.}
\let\thefootnote\relax\footnotetext{$^{**}$Work conducted while at AWS AI Labs.}

\end{abstract}

\begin{figure}[t]
    \centering
    \includegraphics[width=.9\linewidth]{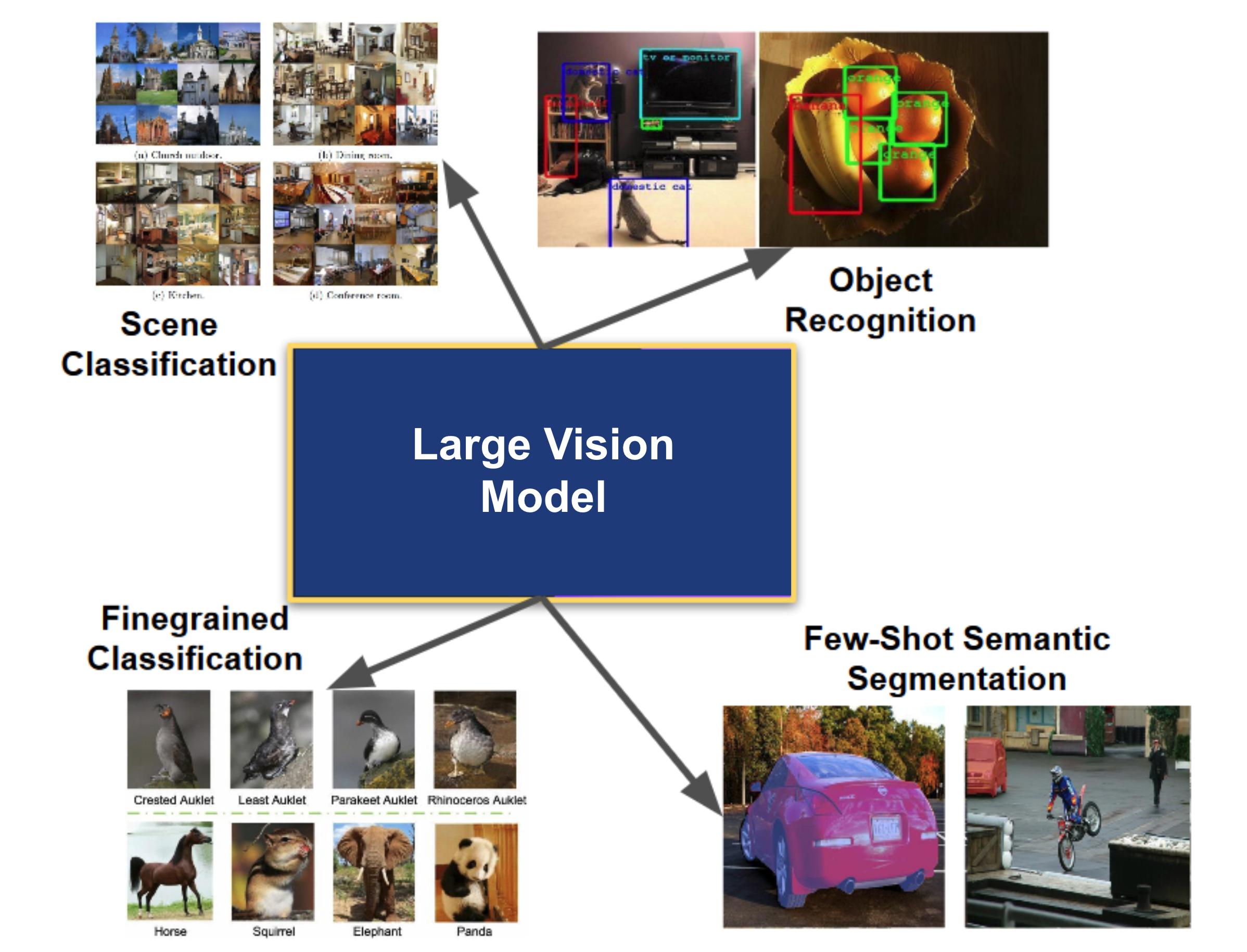}
    \caption{
    In this work we propose EXPRES, a novel adaptation technique for large vision models with the goal of diverse downstream adaptation.
    }
    \label{fig:teaser}
\end{figure}
 \section{Introduction}
\label{sec:intro}
\noindent Scaling up of neural nets in the past few years has steadily improved performance on wide variety of downstream visual tasks. However, model adaptation is often necessary to achieve the best performance in downstream tasks like fine-grained recognition \cite{WelinderEtal2010}, semantic segmentation \cite{DBLP:journals/corr/ChenPK0Y16} or object recognition \cite{8825470}. While traditional techniques like full-model finetuning have become the de-facto approach to  adaptation, they are not well suited for many  scenarios. For example, finetuning is susceptible to catastrophic forgetting \cite{kirkpatrick2017overcoming} as it modifies model parameters without the knowledge of future domains, and potentially losing prior knowledge of current adaptation. Moreover, finetuning all of the model parameters of a large vision model with just a few training examples can lead to poor generalization. This is in contrast to human intelligence that is capable of solving wide variety of downstream tasks with extremely few exemplars. 

\noindent Motivated by the need for better adaptation, parameter efficient techniques like partial-finetuning or adapters\cite{rebuffi2018efficient, zhang2020side} have been developed to constructively adapt large models without significant parameter overhead. While serving as  effective alternatives to finetuning, most parameter efficient techniques have been designed with convolutional architectures in mind. In light of recent works \cite{dosovitskiy2020vit} that demonstrate that Vision Transformers are more suitable for scaling up than CNNs, designing adaptation techniques that exploit the Transformer architecture can be extremely useful. To that end, visual prompt tuning (VPT) \cite{jia2022vpt} has been proposed as a way to constructively adapt transformers by introducing learnable tokens at every layer that interact with the patch and class tokens and are optimized together with a classifier head. While being effective in practice, VPT allows only partial interactions between prompts and the remaining tokens, thus, leveraging only a part of the prompt capacity.
Moreover, it often requires a large number of inserted prompts to achieve optimal performance but that significantly increases the computation costs due to the quadratic computational complexity of the self-attention layer.

\noindent In this work, we explore an alternate design to prompting motivated by the potential for greater prompt capacity. We propose ExPRes, an \textbf{ex}pressive \textbf{p}rompt tuning method with \textbf{res}idual tokens that inherits the strengths of parameter efficient adaptation while significantly improving downstream performance. Our prompt design is inspired by the two key observations - propagation of prompts by multilayered interaction with other tokens is crucial for strong capacity and learnable residual tokens can modulate the propagated prompts to favour task-specific relations (unlike in \cite{jia2022vpt}). We first propagate \textit{shallow} prompts through the encoder that are average pooled at the last layer to yield semantic image-level representations. Shallow prompts by themselves have limited capacity since they cannot specifically modulate token-token relations at higher layers. Therefore to harness the prompts, we add residual tokens to propagated prompts at various layerwise computations of the Transformer encoder including LayerNorm, self-attention and multi-head projection to facilitate layerwise modulation without increasing the number of prompts per layer.  This results in enhanced prompt capacity at almost no additional computational cost.

\noindent We empirically validate the effectiveness of our method on a variety of downstream 
tasks including fine-grained recognition and semantic segmentation. Our use of additional learnable parameters in the form of residual and shallow prompts allows the retention of prior knowledge in the form of frozen encoder weights while being extremely parameter efficient (prompts are $\leq 1 \%$ of the total parameters). Thus, our method is highly suited for real world adaptation that requires information retention at low memory and computational overheads. Additionally, we show that in most cases we require fewer prompts than VPT to achieve the same or better performance, making it more suitable for limited data settings.
Our main contributions can be summarized as follows:
\begin{itemize}
\item We propose a novel prompting technique: EXPRES, that uses a combination of shallow and deep residual prompts to facilitate constructive adaptation to downstream tasks with limited labelled datasets.
\item Our method significantly outperforms full-finetuning based adaptation by $4.6\%$
on VTAB-1k. Moreover, our method outperforms state-of-the-art prompting approach \cite{jia2022vpt} on the same benchmarks with significantly fewer prompts, suggesting that prompt design is crucial to extracting more capacity at a given parameter/computational budget.
\item To the best of our knowledge, we are the first to demonstrate the effectiveness of prompting for diverse applications such as few-shot semantic segmentation. Our method outperforms strong adaptation baselines by $25\%$ and achieves competitive performance with respect to language-assisted segmentation \cite{li2022languagedriven} despite training on significantly less data with dense annotations.
\end{itemize}


\begin{figure*}[t]
    \centering
    \includegraphics[width=.8\textwidth]{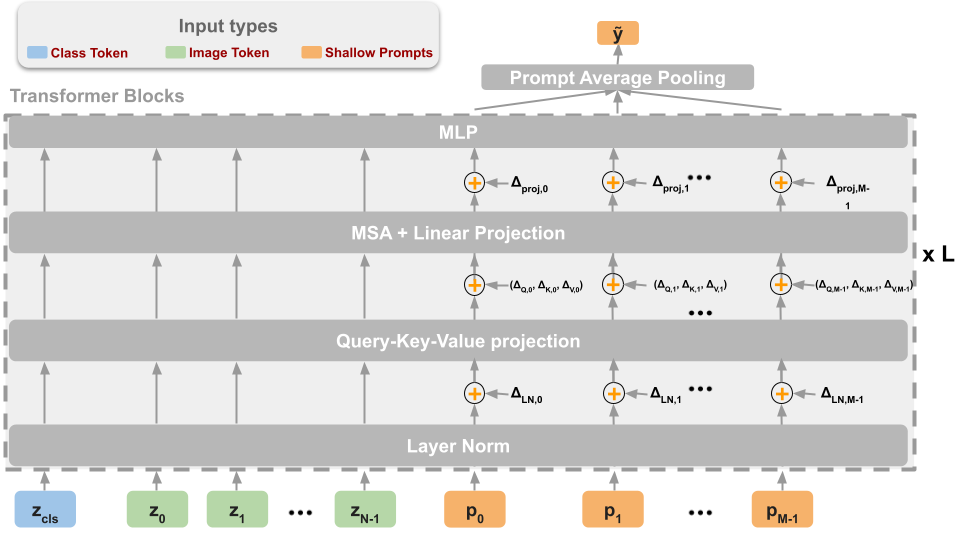}
    \caption{\textbf{\expres{} Architecture in detail:} \expres{} optimizes two types of prompts, shallow prompts (\eg, $p_i$) and residual prompts (\eg, $\nabla_{LN,i}$), to construct task specific representation without updating the pretrained encoder weights. Each residual prompt is a learnable vector that is added to the output of various computations such as Layer Norm, Query-Key-Value projections, and linear projection after the MSA operation.
    }
    \label{fig:main}
\end{figure*}
\section{Related work}
\noindent \textbf{Large Vision Models:} With the advent of Transformer models~\cite{vaswani2017attention} and adoption  to various computer vision tasks, including image classification~\cite{dosovitskiy2020vit,liu2021swin}, object detection~\cite{carion2020end,li2021benchmarking}, semantic and panoptic segmentation~\cite{strudel2021segmenter,zheng2020rethinking,wang2021max}, video understanding~\cite{girdhar2019video,wang2022bevt,feichtenhofer2022masked} and few-shot learning~\cite{doersch2020crosstransformers}, the scale of vision models have increased by orders of magnitude. Typically trained using large labelled data, either unimodal like ImageNet-21K \cite{ILSVRC15} or multimodal, these models demonstrate superior performance on wide variety of visual tasks. Given their superior performance and much larger scale compared to ConvNets, the question of adaptating such models efficiently becomes crucial. Motivated by the need, our work is primarily focussed on adaptation of Vision Transformers.

\noindent \textbf{Transfer Learning} has been extensively studied for vision tasks in the context of ConvNets~\cite{zhuang2020comprehensive} and many techniques have been introduced including side tuning~\cite{zhang2020side}, residual adapter~\cite{rebuffi2017learning}, bias tuning~\cite{cai2020tinytl}, \etc. However, Transformer specific adaptation for visual tasks has received relatively less attention. At the same time, in the NLP domain, the dominance of large-scale pre-trained Transformer-based Large Language Models (LLM)~\cite{devlin-etal-2019-bert,2020t5,brown2020gpt3}, has paved way for many approaches~\cite{he2022towards,guo2020parameter,hu2021lora} that efficiently fine-tune LLMs for different downstream NLP tasks~\cite{wang2018glue,wang2019superglue}. In this work we compare with the most representative methods for fair benchmarking. For example, Adapters~\cite{houlsby2019parameter} insert extra lightweight modules inside each Transformer layer. One adapter module generally consists of a linear down-projection, followed by a nonlinear activation function, and a linear up-projection, together with a residual connection~\cite{pfeiffer2020adapterfusion,pfeiffer2020AdapterHub}. Instead of inserting new modules, \cite{cai2020tinytl} proposed to update the bias term and freeze the rest of backbone parameters when fine-tuning ConvNets. BitFit~\cite{bao2021beit} applied this technique to Transformers and verified its effectiveness on LLM tuning. Through our experiments that we demonstrate that our method, EXPRES provides a more effective way of adapting Transformers  compared to prior approaches.

\noindent \textbf{Prompting:} An  alternative to traditional adaptation methods is prompting~\cite{liu2021pre} - originally proposed as a way of prepending language instruction to the input text so that a pre-trained LLM can ``understand'' the task. Through trial and error selection of appropriate prompts, GPT-3 shows strong generalization to downstream transfer learning tasks even in the few-shot or zero-shot settings~\cite{brown2020gpt3}. This was followed up by other works on better prompt construction~\cite{shin2020autoprompt,jiang2020can}.  Recent works ~\cite{li-liang-2021-prefix,lester-etal-2021-power,liu2021p} propose to treat the prompts as task-specific continuous vectors and directly optimize them via gradients during fine-tuning. Such approaches, named ``Prompt Tuning'' achieve performance comparable to finetuning but with 1000$\times$ less parameters in some cases. Following the success in LLMs, prompt have also been adopted for vision-language models ~\cite{radford2021learning,zhou2021learning,ju2021prompting,yao2021cpt,ge2022domain}. Nonetheless, all the above methods prompt the text encoders and hence are tied to language as input. However, many realistic visual tasks such as dense prediction may not be well aligned with the language modality. Thus, it becomes imperative to develop prompting approaches that can work in the visual modality. 
To that end, recent work on visual prompting \cite{jia2022vpt,conder2022efficient,sandler2022fine, wang2022learning, bahng2022visual} provides encouraging results. In particular, \cite{jia2022vpt} demonstrate that even in the visual domain, adaptation based on continuous prompting can outperform finetuning, especially when the training datasets are small. Our work however shows that current prompting methods do not fully exploit the capacity of prompting for a vision transformer. Through a principled approach to prompting, we derive a more effective prompting technique that achieves state-of-the-art performance at much smaller computational overhead.

\noindent \textbf{Few-Shot Classification:} Few-shot classification has received a lot of attention in recent years. While a variety of approaches \cite{song2022comprehensive} have been proposed, the most successful ones seek to transfer \textit{positive knowledge} either by finetuning~\cite{tian2020rethink, Dhillon2020A, afrasiyabi2020associative} or meta-learning~\cite{Snell2017PrototypicalNF, finn2017model, NIPS2016_6385,sung2018learning, Ravi2017OptimizationAA, Gidaris2018DynamicFV,garcia2017few,fei2021melr,kwon2021repurposing}. Finetuning based few-shot learners can be viewed as specialists that perform well on the target domain   \cite{chen2019closer,tian2020rethink,guo2020broader}, but suffer from catastrophic forgetting \cite{shi2021overcoming} on the base domain. Meta-learning approaches, on the other hand, can be seen as generalists that enjoy complete immunity against forgetting but at the cost of somewhat lower performance in the target domain. In this work, we propose EXPRES as a constructive adaptation technique that is immune to forgetting but at the same time benefits from task specific parameter tuning, thus, leveraging the best of both worlds. One of our key contribution is to demonstrate the transferability of Transformers from classification to dense prediction tasks (semantic segmentation) when adapted with EXPRES.

\noindent \textbf{Few-Shot Semantic Segmentation:} This task was originally proposed in \cite{shaban2017one}. Most works after that follow the metric learning paradigm \cite{dong2018few} with various novelties from improved support-query matching \cite{siam2020weakly,liu2020dynamic,yang2020brinet} to better optimization  \cite{zhu2020self,liu2021learning}, memory modules~\cite{wu2021learning,xie2021few}, graph neural networks~\cite{xie2021scale,wang2020few,zhang2019pyramid}, and more \cite{tian2020differentiable,lu2021simpler, he2021progressive,zhuge2021deep,zhang2021self,liu2021anti,min2021hypercorrelation,li2020fss}.  Some methods generate representative support prototypes with attention mechanism~\cite{zhang2019canet,gairola2020simpropnet}, adaptive prototype learning~\cite{siam2019amp,li2021adaptive,ouyang2020self}, or various prototype generation techniques~\cite{nguyen2019feature,yang2020prototype, wang2019panet,liu2020crnet}. In contrast, our approach directly adapts a model pretrained on image classification to few-shot semantic segmentation without intermediate pretraining on densely-annotated datasets.

\section{Approach}

\subsection{Preliminaries}
Consider an input space $\mathcal{X}$ and a categorical label set $\mathcal{Y}$ where each class is represented via a one-hot encoding. A representation $\mathcal{R} \subset \mathbb{R}^d$ of the input is defined by the composition of an augmentation function $\mathcal{A}:\mathcal{X} \to \mathcal{X}$ and an encoder model $E_\theta : \mathcal{X}\to\mathcal{R}$, parameterized by $\theta$.  The augmentation function is a composition of standard image transformations such as random cropping, horizontal flipping \etc. A general recognition task $\mathcal{T}$ can be defined in terms of a finite training set, $\{(x_i, y_i)\}_{i\in \mathcal{D}_\text{train}}$ and $N_c$ categories such that, $y_i \in [N_c]$. The goal, is to leverage a pre-trained encoder, $E_{\theta^*}$ and the training set to obtain a classifier for the task $\mathcal{T}$.

\subsection{Overview of Vision Transformers (ViT)}
\noindent In a typical ViT, the image, $x \in \mathbb{R}^{h \times w \times 3}$ is uniformly divided into $N$ fixed-sized patches, each of which are projected to a $d$-dimensional embedding and are added a positional embedding, resulting in a patch token $z_n \in \mathbb{R}^{d} $.  Additionally, a class token $z_\text{cls} \in \mathbb{R}^{d}$ is concatenated to the sequence of the patch tokens to form the input, $Z^0 \in \mathbb{R}^{(N+1) \times d}$.  Starting from layer $l=0$, the incoming activations at each layer, $z^{l-1}$ are first normalized using LayerNorm (LN), and then processed by a multi-headed self-attention block (MSA) followed by an MLP block. At the last layer, the resulting class token is normalized to yield the final representation.  The overall computations in a ViT encoder can be summarised as 
\begin{align}
Z^0& = [z_\text{cls}, z_0, \ldots, z_{N-1}] \notag\\
H^{l-1} &= \text{MSA}(\text{LN}(Z^{l-1})) + Z^{l-1} ~~~ l=0,...,L-1  \notag \\
Z^l &=\text{MLP}(\text{LN}(H^{l-1})) + H^{l-1} ~~~ l=0,...,L-1 \notag \\
y &=LN(Z^L_\text{cls}) \notag
\end{align}
where $L$ represents the number of encoder layers.  The multi-headed self-attention mechanism (MSA) abstracts patchwise representation by aggregating the right context at each layer. The context aggregation is facilitated by softmax attention that relies on patch-to-patch similarities and its parameters are governed by the pre-training task objective. During adaptation, however, the representations for downstream task might benefit from aggregating a slightly different context at each layer. Our prompt tuning approach facilitates task specific modulation of this context by augmenting and learning layerwise patch tokens.

\subsection{Expressive Prompt Tuning}

At the input layer of the ViT, we introduce prompt tokens,  $P^0 \in \mathbb{R}^{M\times d} $ which are $M$ parameterized vectors of dimension $d$ concatenated to the input token sequence, $Z^{0}$, of the ViT. Input-level prompts, also referred to as \textit{shallow} prompts, are propagated through the encoder together with the class and patch tokens such that at every layer, each token interacts with every other token through the self-attention layers. The propagated prompts at the last layer are average-pooled to obtain the final representation. The insertion and propagation of the prompts described above can be expressed as 
\begin{align}
\tilde{Z}^0& = [Z^0 || P^0] \label{eq:promptrep} \\
\tilde{H}^{l-1} &= \text{MSA}(\text{LN}(\tilde{Z}^{l-1})) + \tilde{Z}^{l-1} ~~~ l=0,...,L-1 \notag \\ 
\tilde{Z}^l &=\text{MLP}(\text{LN}(\tilde{H}^{l-1})) + \tilde{H}^{l-1} ~~~ l=0,...,L-1 \notag \\
Z^L,  &P^L = \texttt{chunk}(\tilde{Z}^l ) \notag\\
\tilde{y} &= \text{AvgPool}(P^L) \notag \\
y &= LN(\tilde{y})  \notag
\end{align}we use $||$ to denote the concatenation along the sequence axis and \texttt{chunk} to denote the splitting of the propagated sequence of length $N+M+1$ into the $N+1$ tokens and $M$ prompts. Shallow prompts are capable of modeling some of the desired token relations for a task. However, they have limited capacity due to the inability to alter for specific per-layer interactions with the class and image tokens.

To enhance the prompt capacity, we introduce layer-wise residual prompts, $\mathbf{\Delta}^l \in \mathbb{R}^{M \times d}$ that are added to the propagated prompts at various computations within the MSA block for an intermediate layer $l$. This includes the output of attention LayerNorm, query-key-value projections and linear multi-head projection. We summarise the MSA computations at layer $l$ with $N_h$ heads and input $Z$
\begin{align}
Z' &= LN(Z) \notag \\
Q_h &= Z' W_h^Q;~K_h = Z' W_h^K;~V_h = Z' W_h^V \notag \\
\text{g}_h &= \texttt{Att}(Q_h, K_h, V_h) ~~~~~ i=1,...,N_h\label{eq:SA}\\
O &= [\text{g}_1||\ldots||\text{g}_{N_h}]W^\text{proj} \notag 
\end{align}
above $h$ is used to represent the head index of $W^Q, W^K, W^V$ and $W^\text{proj}$ denotes the projection matrix of the MSA block. The residually prompted computations can then be expressed as
\begin{align}
Z' &= LN(Z)+[\bar{\mathbf{0}} || \mathbf{\Delta_{LN}}] \notag\\
\tilde{Q}_h &= Q_h+ [\bar{\mathbf{0}} || \mathbf{\Delta_Q}] \notag\\
\tilde{K}_h &= K_h + [\bar{\mathbf{0}} || \mathbf{\Delta_K}] \notag\\
\tilde{V}_h &= V_h + [\bar{\mathbf{0}} || \mathbf{\Delta_V}] \notag\\
\tilde{O} &= O + [\bar{\mathbf{0}} || \mathbf{\Delta_{proj}}] \notag
\end{align}
where the concatenation of $(N+1) \times d$ dimensional zero matrix, $\bar{\mathbf{0}}$ with the residual-prompts, signifies that the residuals $\mathbf{\Delta}$  are added only at the propagated prompt positions and not to the image patch or class token positions. The above residual computations occur at every layer with independently learned residual prompts, and we drop the layer indices in the equations for brevity. The overall EXPRES architecture is visualized in Figure \ref{fig:main}.

\subsection{Interpreting Residual Prompting}
\noindent We take a closer look at residual prompting for self-attention and interpret it's functionality. The $\texttt{Att}$ operation in Eq. \eqref{eq:SA} facilitates weighted aggregation over all ``value'' tokens where the weights are computed using the ``query'' and a ``key'' tokens as $w_{ij} \propto \text{exp}\left( \frac{q_i^Tk_j}{\sqrt{d/N_h}}\right)$. When prompted with residual tokens, the expression for the weights that aggregate information for a patch token $q_i$ can be factored as,
 \begin{align}
\tilde{w}_{ij} &\propto\text{exp}\left(\frac{q_i^T(k_j + \mathbf{\Delta}_{K,j})}{\sqrt{d/N_h}}\right) \notag \\
&= \text{exp}\left(\frac{q_i^Tk_i }{\sqrt{d/N_h}}\right) \text{exp}\left(\frac{q_i^T\mathbf{\Delta}_{K,j}}{\sqrt{d/N_h}}\right) \label{eq:reweigh} \\
&=w_{ij}*\alpha_{ij}. \notag
\end{align}
Based on Eq. \eqref{eq:reweigh}, residual prompts facilitate task-specific reweighting of the attention weights independently at every layer, allowing the modulation of context aggregated per patch token.
Such layerwise modulation can lead to better adaptation of the final representation compared to shallow prompting that restricts the modulation to the input layer.
Moreover, the two-way interaction between prompts and patch tokens leads to greater flexibility than other forms of multilayered prompting \cite{jia2022vpt,MemFT} that allow for only partial interaction. For instance, \cite{jia2022vpt} restrict the prompts to act only as keys and never as queries.

Residual prompt based attention reweighting is also interesting from parameter efficiency perspective as it circumvents the need for updating the attention weight matrices (done in finetuning approaches) to achieve the same goal of task specific adaptation. Specifically, each layer of a ViT consists of three attention weight matrices, each with $d\times d$ dimensions, resulting in $\mathcal{O}(d^2)$ learnable parameters. In contrast, each prompt is $d$ dimensional so, only $\mathcal{O}(d)$ parameters need to be adapted even with $M$ prompts, where $M << d$. We empirically validate that the high capacity as well as parameter efficiency of our prompting approach is crucial for achieving good adaptation performance in limited labelled data settings. 

 \subsection{Learning the Prompts}
 
\begin{table*}[h]
    \centering
    \resizebox{1
    \textwidth}{!}{
\begin{tabular}{ccccccccc|ccccc|ccccccccc}
\toprule
 &  \multicolumn{8}{c}{natural}  &     \multicolumn{5}{c}{specialized}    &\multicolumn{9}{c}{structured}   \\
\midrule
        & \rotatebox{90}{\bf{CIFAR-100}}
  &\rotatebox{90}{\bf{Caltech101} }
  &\rotatebox{90}{\bf{DTD} }
  &\rotatebox{90}{\bf{Flowers102} }
  &\rotatebox{90}{\bf{Pets} }
  &\rotatebox{90}{\bf{SVHN} }
  &\rotatebox{90}{\bf{Sun397} }
  &\rotatebox{90}{\bf{Mean}}
  &\rotatebox{90}{\bf{Patch Camelyon} }
  &\rotatebox{90}{\bf{EuroSAT} }
  &\rotatebox{90}{\bf{Resisc45} }
  &\rotatebox{90}{\bf{Retinopathy} }
  &\rotatebox{90}{\bf{Mean}}
  &\rotatebox{90}{\bf{Clevr/count} }
  &\rotatebox{90}{\bf{Clevr/distance} }
  &\rotatebox{90}{\bf{DMLab}}
  &\rotatebox{90}{\bf{KITTI/distance} }
  &\rotatebox{90}{\bf{dSprites/location} }
  &\rotatebox{90}{\bf{dSprites/orientation} }
  &\rotatebox{90}{\bf{SmallNORB/azimuth} }
  &\rotatebox{90}{\bf{SmallNORB/elevation} }
  &\rotatebox{90}{\bf{Mean}}  \\
\midrule
Linear &63.4 &85.0 &63.2 &97.0 &86.3 &36.6 &51.0 &68.93  &78.5 &87.5 &68.6 &74.0 &77.16  &34.3 &30.6 &33.2 &55.4 &12.5 &20.0 &9.6 &19.2 &26.84 
\\

MLP-2 &63.2 &84.8 &60.5 &97.6 &85.9 &34.1 &47.8 &67.70  &74.3 &88.8 &67.1 &73.2 &75.86  &45.2 &31.6 &31.8 &55.7 &30.9 &24.6 &16.6 &23.3 &32.47 
\\
MLP-3 &63.8 &84.7 &62.3 &97.4 &84.7 &32.5 &49.2 &67.80  &77.0 &88.0 &70.2 &56.1 &72.83  &47.8 &32.8 &32.3 &58.1 &12.9 &21.2 &15.2 &24.8 &30.62 
\\
MLP-5 &59.3 &84.4 &59.9 &96.1 &84.4 &30.9 &46.8 &65.98  &73.7 &87.2 &64.8 &71.5 &74.31  &50.8 &32.3 &31.5 &56.4 &7.5 &20.8 &14.4 &20.4 &29.23 
\\
MLP-9 &53.1 &80.5 &53.9 &95.1 &82.6 &24.4 &43.7 &61.90  &78.5 &83.0 &60.2 &72.3 &73.49  &47.5 &27.9 &28.9 &54.0 &6.2 &17.7 &10.8 &16.2 &26.15 
\\
\midrule
Sidetune \cite{zhang2020side}  &60.7 &60.8 &53.6 &95.5 &66.7 &34.9 &35.3 &58.21  &58.5 &87.7 &65.2 &61.0 &68.12  &27.6 &22.6 &31.3 &51.7 &8.2 &14.4 &9.8 &21.8 &23.41 
\\
Biastune \cite{cai2020tinytl} &72.8 &87.0 &59.2 &97.5 &85.3 &59.9 &51.4 &73.30 &78.7 &91.6 &72.9 &69.8 &78.25  &61.5 &55.6 &32.4 &55.9 &66.6 &40.0 &15.7 &25.1 &44.09 
\\
Adapter-256 &74.1 &86.1 &63.2 &97.7 &87.0 &34.6 &50.8 &70.50  &76.3 &88.0 &73.1 &70.5 &76.98  &45.7 &37.4 &31.2 &53.2 &30.3 &25.4 &13.8 &22.1 &32.39 
\\
Adapter-64 &74.2 &85.8 &62.7 &97.6 &87.2 &36.3 &50.9 &70.65  &76.3 &87.5 &73.7 &70.9 &77.10  &42.9 &39.9 &30.4 &54.5 &31.9 &25.6 &13.5 &21.4 &32.51 
\\
Adapter-8 &74.2 &85.7 &62.7 &97.8 &87.2 &36.4 &50.7 &70.67  &76.9 &89.2 &73.5 &71.6 &77.80  &45.2 &41.8 &31.1 &56.4 &30.4 &24.6 &13.2 &22.0 &33.09 \\
Partial-1 &66.8 &85.9 &62.5 &97.3 &85.5 &37.6 &50.6 &69.44  &78.6 &89.8 &72.5 &73.3 &78.53  &41.5 &34.3 &33.9 &61.0 &31.3 &32.8 &16.3 &22.4 &34.17 
\\
FT-all &68.9 &87.7 &64.3 &97.2 &86.9 &\bfseries87.4 &38.8 &75.88 &79.7 &95.7 &\bfseries84.2 &73.9 &83.36 &56.3 &58.6 &41.7 &65.5 &57.5 &46.7 &25.7 &29.1 &47.64 
\\
\midrule
VPT-shallow \cite{jia2022vpt} &77.7 &86.9 &62.6 &97.5 &87.3 &74.5 &51.2 &76.81  &78.2 &92.0 &75.6 &72.9 &79.66  &50.5 &58.6 &40.5 &67.1 &68.7 &36.1 &20.2 &34.1 &46.98 \\
 VPT-deep \cite{jia2022vpt} &\bfseries78.8 &\bfseries90.8 &65.8 &98.0 &88.3 &78.1 &49.6 &78.48  &81.8 &96.1 &83.4 &68.4 &82.43   &\bfseries68.5 &60.0 &\bfseries46.5 &72.8 &73.6 &47.9 &\bfseries32.9 &\bfseries37.8 &54.98 \\
 \midrule
 EXPRES (\textit{ours})&78.0&89.6&\bfseries68.8&\bfseries98.7&\bfseries88.9&81.9&\bfseries51.9&\bfseries79.7&\bfseries84.8&\bfseries96.2&80.9&\bfseries74.2&\bfseries84.0&66.5&\bfseries60.4&\bfseries46.5&\bfseries77.6&\bfseries78.0&\bfseries49.5&26.1&35.3&\bfseries55.0\\
 \bottomrule
\end{tabular}    }
    \caption{\textbf{VTAB-1k benchmark:} Per task adaptation results with ViT-B/16 model pretrained on ImageNet-21k..}
    \label{tab:vtab}
\end{table*}

In this work, we are mainly interested in two types of downstream tasks - image classification and semantic segmentation. To train the prompts for classification, we optimize a standard cross entropy loss with respect to the representation, $y$ in Eq. \eqref{eq:promptrep} and the corresponding ground-truth label, $y*$. In the case of semantic segmentation, we adapt the model as well as the objective to perform dense predictions. At the final layer of the encoder, we extract the keys corresponding to the patch tokens and pass them through the classifier. The sequence outputs are then reshaped into a 2d map and resized to original image resolution using bilinear interpolation, resulting in a pixel wise prediction. Finally, to optimize the prompts with the classifier head, we use a dense cross entropy loss as follows
\begin{align}
\{p_m^*\}, \{\mathbf{\Delta^*}\} = \argmin\limits_{\{p_m\}, \{\mathbf{\Delta}\}, C}  \sum_{j\in I_\text{ctxt}} \sum_{h,w}  L_\text{CE}(y_{jhw}, y_{jhw}^*) \notag
\end{align}
where,  $h,w$ are used to index the spatial positions at the resolution, $H\times W$ of the input image.

\section{Experiments}

\begin{table*}[t]
    \centering
    \scriptsize
    \begin{tabular}{cccc|cccc}
\toprule
Method & Im-Pre & Dense-Pre & Backbone &  $5^0$ & $5^1$ & $5^2$  & $5^3$\\
\midrule
 SPNET\cite{8953827}&In-1k&\checkmark&ResNet-101&23.8	&17.0&14.1&18.3\\
 ZS3Net\cite{NEURIPS2019_0266e33d}&In-1k&\checkmark&ResNet-101&40.8&39.4&39.3&33.6\\
 LSEG\cite{li2022languagedriven}&In-1k&\checkmark&ViT-L/16&\textbf{61.3}&\textbf{63.6}&43.1&41.0\\
\midrule
Linear&In-21k&-&ViT-B/16&5.1&43.9&28.1&29.1\\
FT-all&In-21k&-&ViT-B/16&18.3&31.9&27.9&23.2\\
Biastune&In-21k&-&ViT-B/16&2.4&39.2&13.9&20.1\\
\midrule

VPT-deep \cite{jia2022vpt}&In-21k&-&ViT-B/16 &33.4&56.2&49.8&47.7\\
\expres{} (\textit{\textbf{Ours}})&In-21k&-&ViT-B/16&41.8&60.2&\bfseries52.4&\bfseries51.4\\
\bottomrule
\end{tabular}
\caption{\textbf{Five-Shot Semantic Segmentation on \pascal{i}:} Per fold adaptation results with ViT-B/16 model pretrained on ImageNet-21k.}
    \label{tab:fss_voc}
\end{table*}

We validate the effectiveness of \expres{} on a variety of benchmarks consisting of wide variety of tasks and dataset sizes. We also analyse the importance of various model components and design decisions. 

\noindent\textbf{Datasets: }To evaluate \expres{}, we use two different benchmarks, \vtab{} \cite{zhai2019largescale} and FGVC\cite{jia2022vpt}.  The \textbf{\vtab{}} benchmark consists of $19$ different visual classification tasks categorized under three groups: \textit{Natural} - tasks with natural images captured with standard cameras; \textit{Specialized} - tasks with images captured under specialized settings (medical and satellite imagery); and \textit{Structured} - tasks that requires understanding scene geometry, like object distance. Each task-specific dataset contains $1000$ training examples with varying number of samples per class, depending on the number of classes. For validation purposes and hyper-parameter selection, we use a $800-200$ split of the training set and then train on all $1000$ examples for final results, which are based on evaluation on the entire test set. The \textbf{FGVC} benchmark consists of the finegrained datasets including CUB \cite{WelinderEtal2010}, Oxford Flowers \cite{4756141}, Stanford Dogs \cite{KhoslaYaoJayadevaprakashFeiFei_FGVC2011} and Stanford Cars \cite{10.5555/3298023.3298221}. In conjunction with \vtab{}, we use the FGVC datasets to conduct key ablation studies for \expres{}. A random $90-10$  split of each dataset is used for hyperparameter selection. For few shot segmentation, we use the standard  \textbf{\pascal{i}} \cite{OneShot} benchmark that was created from PASCAL VOC 2012 \cite{Everingham2014ThePV} with extra mask annotations for 20 object classes, evenly divided into $4$ folds: $\{5^i: i \in \{0, 1, 2, 3\}\}$. Following prior works for evaluation scheme, we randomly sample $1000$ episodes per fold and report the average performance over all episodes of the corresponding fold. However, unlike prior works, we tune our \expres{} model on few-shot training examples and cross-validate the hyperparameters (\eg number of prompts, $M$) on a validation set with fold-exclusive categories. In all our experiments,  we use a reasonably small budget for hyperparameters (see supplementary) following recent studies \cite{oliver2018realistic} that highlight the likelihood of overoptimistic results in limited-labelled data settings due to  excessive hyperparameter tuning on large validation sets. \\
\textbf{Implementation Details: } In all our experiments, we use a fixed encoder, ViT-B/16 pretrained on ImageNet-22K \cite{ILSVRC15}. This model is effective on wide variety of tasks and allows direct comparison with prior works. For each downstream task, we train for a total $100$ epochs with an initial warmup of $10$ epochs. We use AdamW as our default optimizer with a suitable learning rate and fixed weight decay of $1e-4$. For \vtab{} and FGVC experiments, we use a fixed batch size of $64$. For segmentation experiments, since overall training set sizes are extremely small ($\leq 10$ images in total, we use the entire training set per batch. We use input image resolution of $224\times 224$ for classification tasks and $384\times 384$ for semantic segmentation tasks as dense prediction usually benefits from larger resolution. Since, ViT-B/16 is pretrained on standard  $224\times 224$ resolution images, we use interpolated positional embeddings to accommodate larger resolutions in the case of segmentation. Finally, we use standard data augmentations for classification benchmarks \ie, \texttt{Resize} $\to$ \texttt{Random-Crop} $\to$ \texttt{Horizontal-Flip} during training and \texttt{Resize} $\to$ \texttt{Center-Crop} during evaluation while for segmentation we only use \texttt{Resize}.

\noindent\textbf{Evaluation Metrics} For classification experiments, we use accuracy as our performance metric. For evaluating segmentation masks, we use mean intersection over union (mIoU) that averages over the intersection over union curves per class.

\subsection{Main Results}

\noindent We compare \expres{} with a number of commonly used adaptation techniques on \vtab{}. The adaptation methods can be categorized as \textit{head-oriented}, \textit{backbone-oriented} and \textit{prompt-based}. Under the first category,   \textit{Linear} only optimizes a linear classifier for the downstream task while \textit{MLP-k} uses a $k$-layer multilayer perceptron (MLP) as the classifier head. As an example of the second category, \textit{Sidetune} \cite{zhang2020side} uses features that are linearly interpolated between pretrained features and features from a ``side'' network trained on downstream data.  While \textit{Biastune} \cite{cai2020tinytl} adapts only the bias terms of an otherwise frozen backbone. \textit{Adapter-d} \cite{houlsby2019parameter,pfeiffer2020adapterfusion,pfeiffer2020AdapterHub} introduces lightweight MLP modules inserted between Transformer layers. \textit{Partial-k} finetunes the last \textit{k} layers while keeping the rest of the backbone frozen, finally \textit{FT-all} finetunes all the layers. 
\textit{VPT-shallow} \cite{jia2022vpt} additionally optimizes a linear classifier with the learnable input prompts propagated through the encoder. \textit{VPT-deep} \cite{jia2022vpt} introduces more capacity by replacing the propagated prompts with a new set of learnable prompts at each layer. 

\noindent\textbf{\vtab{} Results (Table \ref{tab:vtab}):} Across all three splits of \vtab{}, our method significantly outperforms the best \textit{head-oriented} techniques: $~+11\%$ (\textit{natural}), $~+7\%$ (\textit{specialized}) and $~+23\%$ (\textit{structured}). Similar trends hold even when comparing with the more powerful class of \textit{backbone-oriented} techniques. While \textit{FT-all} has been widely adopted as an effective technique for most adaptation scenarios, our method consistently outperforms it by a significant margin of $~+4\%$ (\textit{natural}), $~+1\%$ (\textit{specialized}) and $~+7\%$ (\textit{structured}). Most interestingly our method even outperforms other powerful prompting techniques like \textit{VPT-deep} on $12$ (out of $19$) datasets by a margin of about  $~+1\%$ (\textit{natural}), $~+2\%$ (\textit{specialized}) and $~+0.02\%$ (\textit{structured}). The performance gains are particularly impressive when considered from the perspective of computation vs performance tradeoff. Compared to \textit{VPT-deep} that uses $53$ prompts for \textit{natural} and $108$ prompts for structures, our method only requires $10$ and $29$ prompts for the respective splits. A more detailed summary is provided in the supplementary. 

\begin{figure}[ht]
    \centering
\includegraphics[width=0.8\columnwidth]{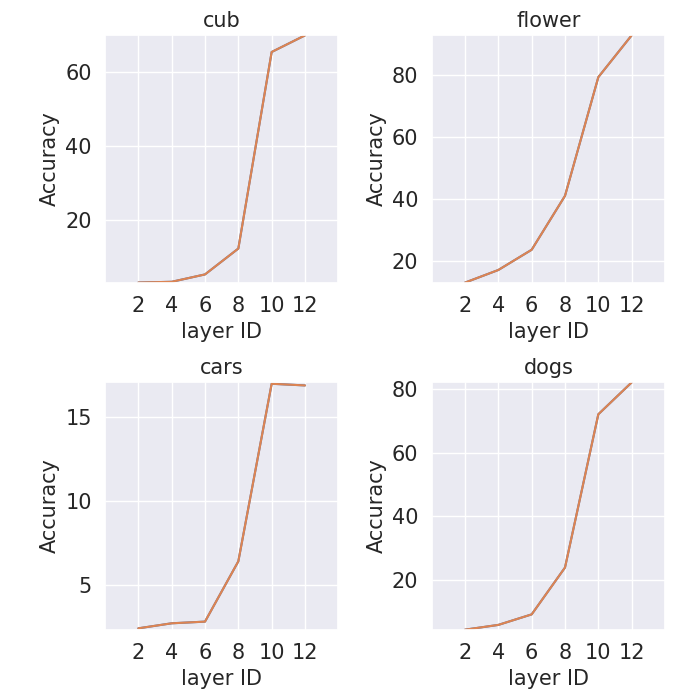}
    \caption{\textbf{Prompt Propagation:} Effect of propagating prompts with modulation upto a layer, $l=\{2, \ldots, 12\}$ of the ViT-B/16 encoder with total $12$ layers. The datasets are sampled from the FGVC benchmark.}
    \label{fig:propag}
\end{figure}
\begin{figure*}[ht]
    \centering
\includegraphics[width=0.8\linewidth]{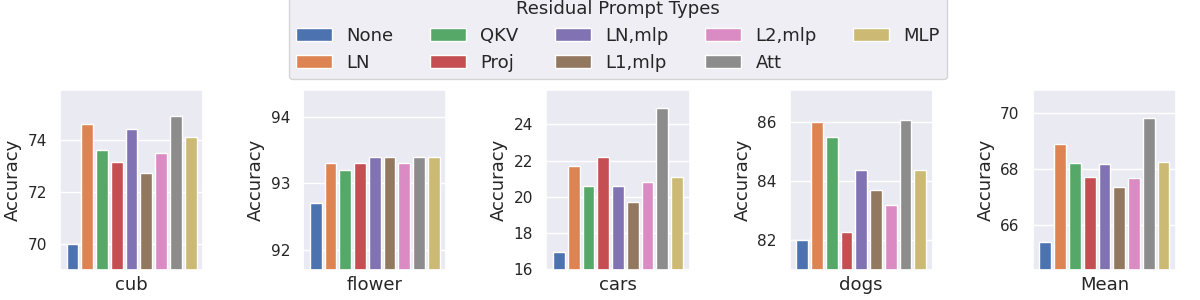}
    \caption{\textbf{Residual Prompt Types:} Evaluating the importance of different type of residual prompts on FGVC datasets.}
    \label{fig:ab_ptype}
\end{figure*}

\begin{figure}[ht]
\centering
\includegraphics[width=0.8\columnwidth]{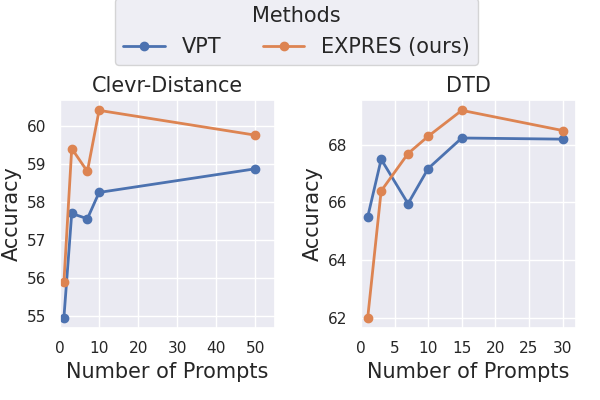}
    \caption{\textbf{Computational Efficency Plots:} Comparing accuracy and number of prompts for VPT and \expres{}}
    \label{fig:compef}
\end{figure}

\subsection{Few-Shot Semantic Segmentation Without Dense Pretraining}
\label{sec:fss}

\noindent We test the efficacy of \expres{} on a novel adaptation setup where the backbone, pretrained on classification tasks, is directly adapted for few-shot semantic segmentation task without additional training on large densely-annotated datasets. This is in contrast with most works that follow a two-stage pretraining procedure \ie, training on ImageNet-1k with image level labels followed by meta-learning on densely annotated datasets constructed from the train folds of \pascal{i}. In contrast, we only perform the first stage pretraining on a sufficiently diverse dataset (ImageNet-21k in our case) with no meta-learning stage. During evaluation, each few-shot episode randomly samples a target image with a segmentation mask that assigns a label $1$ to the pixels corresponding to one of the object categories in that image and a label $0$ to remaining pixels (background). From the same object category, a set of five images are randomly sampled to form the training set where the mask of each image is annotated in the same way as the target. Our method as well as all baselines that perform meta-learning are optimized given a context set to yield a binary classifier for the target image. For dense representations, we use the last layer \textit{keys} of the MSA-block as we found them to be more accurate than the typical MLP-block output. We observe that our method consistently outperforms the baselines (by $25\%$) as well as other prompt techniques such as VPT-deep (by $5\%$). Surprisingly, our method even outperforms \cite{li2022languagedriven} that leverages densely annotated datasets (training data per fold) with large language models for a second stage of pretraining. Specifically, on 2/4 folds, \expres{} outperforms \cite{li2022languagedriven} by $10\%$ despite training on a significantly smaller densely-annotated dataset (five images). These results are particularly significant as they suggest that highly competitive results can be achieved with models pretrained on image classification. Consequently, results may be further improved by scaling up the image-level annotated datasets which are cheaper to scale than the densely annotated datasets. In the supplementary, we provide visualizations of the segmentation mask predictions by our method as well as Linear, FT-all and VPT-deep methods.

\subsection{Ablation Studies}
\label{subsec:ablate}
We conduct extensive ablations to evaluate key design decisions used to develop \expres{} such as feature construction, residual prompting, number of prompts \emph{etc.} For all ablations, we use the same ViT-B/16 backbone. When using FGVC datasets, we use only $10\%$ of each dataset with official splits provided in \cite{jia2022vpt}. 

\noindent\textbf{Prompt Propagation:} We evaluate the importance of propagating prompts through the ViT-B/16 encoder with layerwise modulation. Specifically,  upto a layer $l$, we allow all tokens (patch, class and prompt) to attend to each other at every layer. Beyond $l$,  prompts are not allowed to interact with other tokens at all; they are simply projected by the value heads of MSA-block followed by MLP processing at every layer. In our experiments, we fix the number of prompts to $10$ irrespective of the dataset. From Figure \ref{fig:propag}, we observe that downstream performance depends directly on the extent of prompt propagation with modulation: accuracy improves as more layers allow prompts to interact with other tokens. These results motivate our method, \expres{}, that facilitates finegrained layerwise modulation via residual tokens.

\noindent \textbf{Residual Prompt Type:} In Figure \ref{fig:ab_ptype} we evaluate the importance of different residual prompt types (one-at-a-time) including attention, LayerNorm (\textit{LN}), query-key-value projections (\textit{QKV}) and linear multi-head projection (\textit{Proj}). We also evaluate residual prompting in the MLP block: after LayerNorm (\textit{LN,mlp}), first linear projection (\textit{L1,mlp}) and second linear projection (\textit{L2,mlp}). We evaluate the composite effect of multiple prompt types within a computational block \ie, \textit{Att} (MSA block) and \textit{MLP} (MLP block)
As a baseline, we provide the per-dataset results for shallow prompting, referred to as \textit{None} in the figure. The number of prompts at each layer are fixed at $10$. We observe that, within the MSA block, adding residual prompts to LayerNorm  and  query-key-value projections yields the most improvements ($3.5\%$ and $2.8\%$ on an average respectively) over \textit{None}. Moreover, LayerNorm prompts in the MLP block are also more effective than prompting the two linear layers. Comparing blockwise performance, prompting the MSA block (\textit{Att}) yields significantly better performance (by $1.6\%$) than  prompting the MLP block (\textit{MLP}).
The performance gap between \textit{Att} and \textit{MLP} highlights the importance of directly modulating layerwise interaction between tokens for better adaptation. Consequently, we use \textit{Att} as our default setting for all experiments. 

\noindent \textbf{Computational Efficiency of Prompting}: In prompting techniques, the primary computational overhead arises from the quadratic complexity (in number of tokens)  of the transformer encoder. So to evaluate computation efficiency of prompting, we investigate the rate at which performance improves with number of prompts and provide comparisons between our method and VPT in Figure \ref{fig:compef}. We evaluate on two different datasets sampled from different categories of \vtab{}. For a given accuracy, \eg $58.5\%$ on Clevr-Distance, \expres{} requires an order less prompts than VPT, resulting in 2 orders less computations. Overall \expres{} achieves higher optimal performance with far fewer prompts than VPT.

\section{Conclusion}
In this work we propose a novel prompting technique for adapting large vision models. Our method demonstrates strong performance across variety of downstream tasks with varying dataset sizes. Further, our method 
outperforms commonly used finetuning approach as well as the recently proposed VPT method on standard benchmarks. We also demonstrate diverse adaptation ability of our method from classification to semantic segmentation tasks in the few-shot setting. Lastly, our method is more parameter efficient that existing weight-space and prompt based adaptation techniques. In the future, we plan to extend our method to additional settings including vision-language learning.

{\small
\bibliographystyle{ieee_fullname}
\bibliography{egbib}
}

 \newpage
 \appendix

\section{Hyperparamter Details for Main Results}
\begin{table}[h]
    \centering
    \begin{tabular}{cc}
    \toprule
    Hyperparameter &Range\\
    \midrule
         optimizer& ADAMW\\
         learning-rate, ($lr$)& $\{0.005, 0.001, 0.0005, 0.0001\}$ \\
         weight-decay,  ($wd$)& $\{0.0001,0.001\}$ \\
         epochs & 100 \\
         warmup epochs& 10 \\
         \bottomrule
    \end{tabular}
    \caption{\textbf{Hyperparameter Range}}
    \label{tab:hypam_summ}
\end{table}

\begin{table*}[h]
    \centering
    \resizebox{1
    \textwidth}{!}{
\begin{tabular}{ccccccccc|ccccc|ccccccccc}
\toprule
 &  \multicolumn{8}{c}{Natural}  &     \multicolumn{5}{c}{Specialized}    &\multicolumn{9}{c}{Structured}   \\
\midrule
        & \rotatebox{90}{\bf{CIFAR-100}}
  &\rotatebox{90}{\bf{Caltech101} }
  &\rotatebox{90}{\bf{DTD} }
  &\rotatebox{90}{\bf{Flowers102} }
  &\rotatebox{90}{\bf{Pets} }
  &\rotatebox{90}{\bf{SVHN} }
  &\rotatebox{90}{\bf{Sun397} }
  &\rotatebox{90}{\bf{Mean}}
  &\rotatebox{90}{\bf{Patch Camelyon} }
  &\rotatebox{90}{\bf{EuroSAT} }
  &\rotatebox{90}{\bf{Resisc45} }
  &\rotatebox{90}{\bf{Retinopathy} }
  &\rotatebox{90}{\bf{Mean}}
  &\rotatebox{90}{\bf{Clevr/count} }
  &\rotatebox{90}{\bf{Clevr/distance} }
  &\rotatebox{90}{\bf{DMLab}}
  &\rotatebox{90}{\bf{KITTI/distance} }
  &\rotatebox{90}{\bf{dSprites/location} }
  &\rotatebox{90}{\bf{dSprites/orientation} }
  &\rotatebox{90}{\bf{SmallNORB/azimuth} }
  &\rotatebox{90}{\bf{SmallNORB/elevation} }
  &\rotatebox{90}{\bf{Mean}}  \\
\midrule
VPT-shallow &77.7 &86.9 &62.6 &97.5 &87.3 &74.5 &51.2 &76.81  &78.2 &92.0 &75.6 &72.9 &79.66  &50.5 &58.6 &40.5 &67.1 &68.7 &36.1 &20.2 &34.1 &46.98 \\
num. prompts ($M$)&100 &5 &1 &200 &50 &200 &1 &79.4 &5 &50 &50 &10 &28.7 &100 &200 &100 &100 &100 &100 &200 &200 &137.5\\
 \midrule
 VPT-deep &\bfseries78.8 &\bfseries90.8 &65.8 &98.0 &88.3 &78.1 &49.6 &78.48  &81.8 &96.1 &83.4 &68.4 &82.43   &\bfseries68.5 &60.0 &\bfseries46.5 &72.8 &73.6 &47.9 &\bfseries32.9 &\bfseries37.8 &54.98 \\
 num. prompts ($M$)&10 &10 &10 &1 &1 &50 &5 &12.4 &100 &100 &10 &1 &52.8 &50 &200 &100 &50 &10 &50 &200 &200 &107.5
\\
 \midrule
 EXPRES (\textit{ours})&78.0&89.6&\bfseries68.8&\bfseries98.7&\bfseries88.9&81.9&\bfseries51.9&\bfseries79.7&\bfseries84.8&\bfseries96.2&80.9&\bfseries74.2&\bfseries84.0&66.5&\bfseries60.4&\bfseries46.5&\bfseries77.6&\bfseries78.0&\bfseries49.5&26.1&35.3&\bfseries55.0\\
 num. prompts ($M$)&30&10&15&10&5&10&10&12.9&30&10&10&10&15.0&100&10&10&10&30&30&10&30&28.75\\
 \bottomrule
\end{tabular}    }
    \caption{\textbf{Extended \vtab{} results:} Comparing \expres{} with VPT-shallow and VPT-deep (VPT) with optimal number of prompts per \vtab{} task. The highest accuracies are highlighted per task.}
    \label{tab:vtab_hypam}
\end{table*}

\noindent The hyperparameters and their ranges used in our main experiments are tabulated in Table \ref{tab:hypam_summ}. In Table \ref{tab:vtab_hypam}, we report the original results of Table \ref{tab:vtab} with optimal number of prompts per task.When comparing to other prompting techniques like VPT, we use the official implementation\footnote{\url{https://github.com/KMnP/vpt}}.  We observe that in many tasks (like SVHN, Patch-Cam., Clevr/distance, Kitti/dist., dsprites/orient.), \expres{} outperforms VPT-deep with significantly fewer prompts. 

\section{Effect of Hyperparameters}
\label{sec:hypam_suppl}

\begin{figure}[h]
    \centering

\includegraphics[width=0.9\columnwidth]{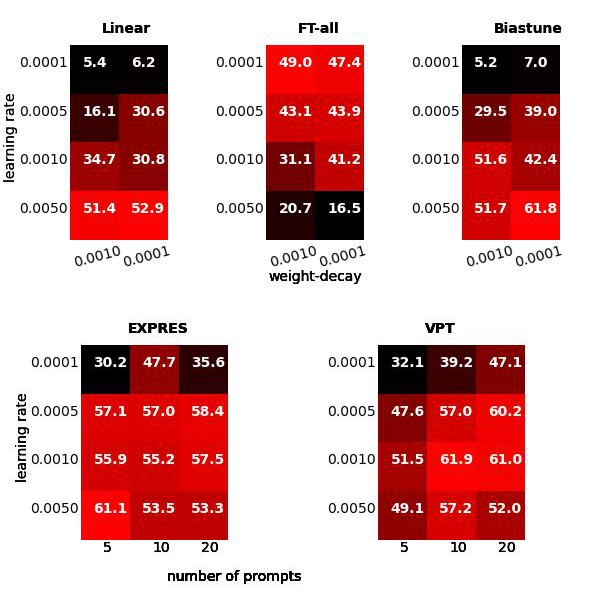}
\caption{\textbf{Effect of hyperparameters for Five-Shot Semantic Segmentation on \pascal{0}}}
    \label{fig:hypam_voc}
\end{figure}

 \begin{figure}[h]
     \centering
     \includegraphics[width=0.9\columnwidth]{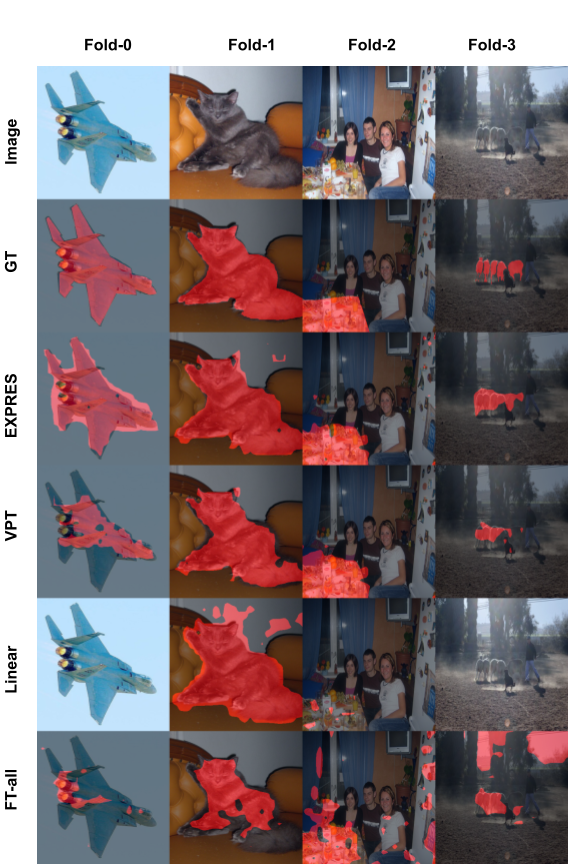}
     \caption{Five-Shot Predictions on \pascal{i}}
     \label{fig:viseg}
 \end{figure}

\noindent We study the effect of hyperparameters on various baselines and \expres{} used for five-shot segmentation. To create the validation set for the $i^\text{th}$ fold, we sample $100$ random tasks from the corresponding train-split that consists of fold exclusive categories, $\{5^j|j\neq i\}$ and evaluate the average accuracy over these tasks. For each baseline method, we test the following learning rates: $\{0.0001, 0.0005, 0.001, 0.005\}$. For prompt based methods, we fix the weight decay to $0.0001$ and test following number of prompts: $\{5,10,20\}$. For rest of the methods (non-prompt-based), we vary the weight decay in the set, $\{0.001,0.0001\}$. In Figure \ref{fig:hypam_voc}, ADAMW is chosen as the optimizer and fold-$5^0$ as the fold for analysis. For other hyperparameters like number of epochs and warmup epochs, fixed values of $100$ and $10$ respectively worked well. We use the above hyperparameters validation to pick the optimal values for final evaluation. In Figure \ref{fig:viseg}, we visualize the predictions for \expres{} and compare it to various baselines. \expres{} tends to produce more complete segmentation masks than others across different folds.

\section{Additional Ablations on \vtab{}}
\begin{figure}[h]
    \centering
    \scriptsize
\includegraphics[width=0.9\columnwidth]{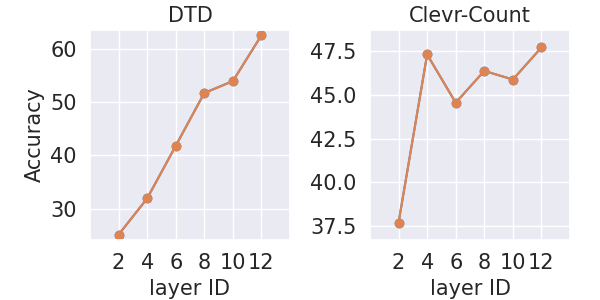}
    \caption{\textbf{Prompt propagation (additional ablations):} Effect of propagating prompts with modulation upto a layer, $l=\{2, \ldots, 12\}$ of the ViT-B/16 encoder with total $12$ layers. The datasets are sampled from the \vtab{} benchmark.}
    \label{fig:propag_suppl}
\end{figure}

\begin{figure*}[t]
    \centering

\includegraphics[width=0.9\linewidth]{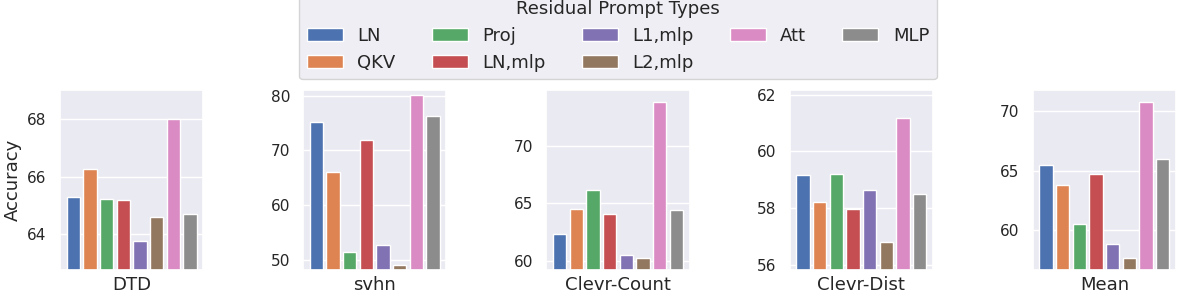}
    \caption{\textbf{Residual Prompt Types (additional ablations):} Evaluating the importance of different type of residual prompts on \vtab{} datasets.}
    \label{fig:ab_ptype_suppl}
\end{figure*}

\noindent We demonstrate the importance of \textbf{prompt propagation} and \textbf{ residual prompts} on additional tasks (\textit{DTD} and \textit{Clevr-Count}) from the \vtab{} benchmark. In all ablations, we use optimal learning rates and weight decays with $M=15$ for \textit{DTD} task and $M=10$ for \textit{Clevr-Count} task.   In Figure \ref{fig:propag_suppl}, we repeat the ablation of propagating shallow prompts without residual prompting for additional datasets. We observe similar trends as  in Figure \ref{fig:propag}- downstream performance depends directly on the extent of prompt 
propagation with modulation \ie, interaction with other tokens via self-attention. We also observe that for \textit{Clevr-Count}, the performance quickly rises upto layer $4$, then plateaus. Thus, the exact performance trend with increasing propagation through the layers varies slightly with the downstream task.

\noindent In Figure \ref{fig:ab_ptype_suppl}, we repeat the ablation that delineates the importance of each type of residual prompt for additional datasets. We observe that trends are similar to  Figure \ref{fig:ab_ptype}. Within each block (\textit{Att} and \textit{MLP}), layer norm residual prompts yield notable improvements in most tasks. Within MSA block (\textit{Att}), the \textit{QKV} prompts are dominant over \textit{Proj} prompts for \textit{natural} tasks (DTD, svhn) of \vtab{} while the trend reverses for \textit{structured} tasks (Clevr-Dist, Clevr-Count). Most importantly, when comparing blockwise performance, prompting the MSA block (\textit{Att}) consistently outperforms prompting the MLP block (\textit{MLP}), reinforcing the conclusions in \S \ref{subsec:ablate}.

\section{Effect of Prompt Initiation Layer}

 \begin{figure}[t]
       \includegraphics[width=1.1\columnwidth]{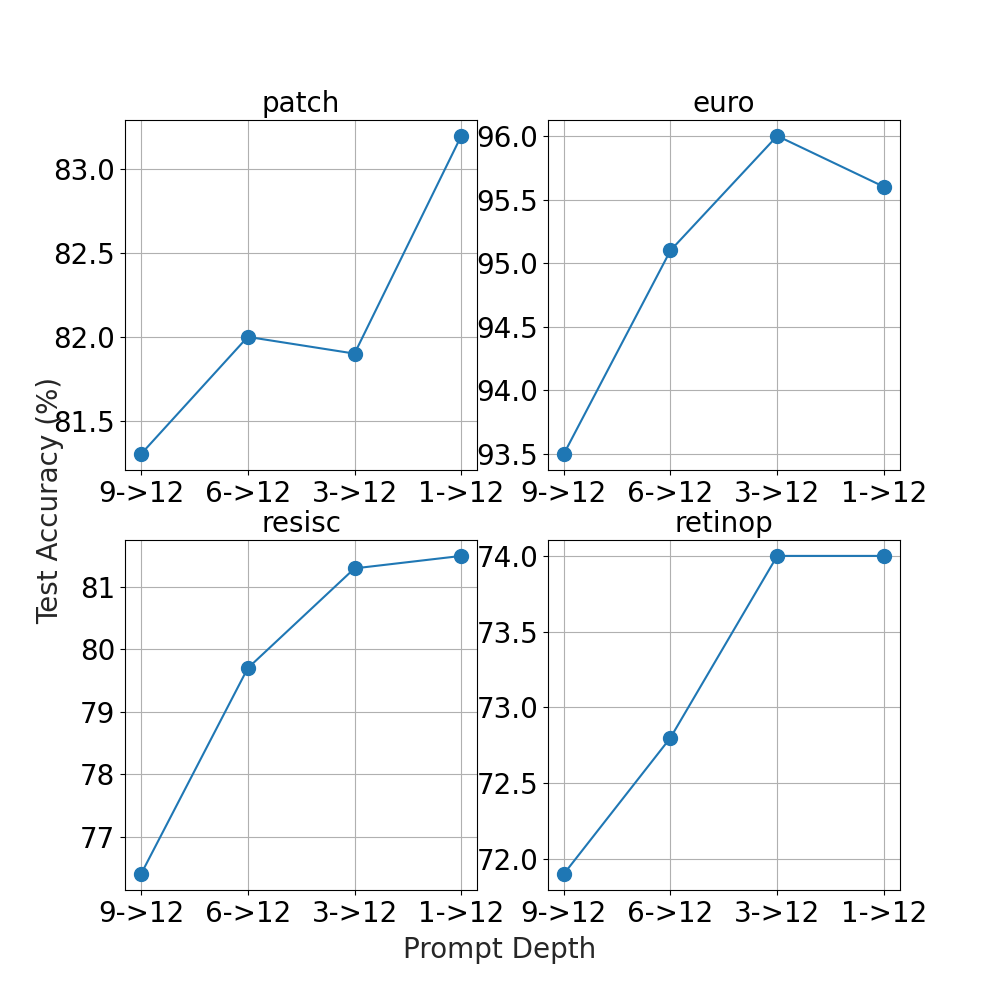}
        \caption{\textbf{Starting \expres{} prompting at specific layers: } $l_1 \to l_2$ means prompting starts at $l_1$ (closer to input) and ends at $l_2$.  number of prompts is fixed at $10$.}
     \label{fig:prompt_layer}
 \end{figure}
\noindent One of the key design decisions of \expres{} prompting is where to insert the prompts. In Fig. \ref{fig:prompt_layer} we show that initiating \expres{} prompting at early layers is generally a good strategy. Combined with the observations in \ref{fig:propag}, we recommend using prompts at every layer to ensure best performance.

\section{Effect of Architectural Choice}

\begin{table}[h]
\setlength{\tabcolsep}{2pt}
    \centering
\resizebox{0.5\textwidth}{!}{
\begin{tabular}{cccccccc}
\toprule 
&Linear&MLP-3&Partial-1&Biastune&\vpt{}&\vptsh{}&\expres{}\\

\midrule 
VTAB-\textit{Specialized} &80.8&75.2&81.7&80.1&84.5&82.5&\bfseries84.6\\
	 \bottomrule
\end{tabular}  
}
    \caption{\textbf{Comparing various methods for adapting Swin-Base model pretrained on ImageNet-21k}}
    \label{tab:swin}
\end{table}

\noindent\expres{} was designed as a generic prompting technique for Transformer architectures. To demonstrate it's generality, we compare our method with other adaptation techniques with Swin \cite{liu2021swin} transformers as the backbone. In particular, we incorporate \expres{} prompts in Swin-Base architecture, where fixed number (10) of shallow prompts are propagated through all four stages without any modification during patch merging with residual prompts being added at each layer of the SwinTF-blocks. Evaluations on a subset of the \vtab{} dataset in Table \ref{tab:swin} shows that overall our method outperforms various baselines and state-of-the-art methods even when applied to a different variant of transformer architecure.

\section{FGVC results}
\begin{table}
\scriptsize
\begin{center}
\begin{tabular}{
crrrrr
}
\toprule
  &Cub&Flower&Dogs&Cars\\
  \\
\midrule
\fullft{} &87.3 &98.8 &89.4 &\bfseries84.5
\\
\linear{} &85.3 &97.9 &86.2 &51.3
\\
\partialft{}-1 &85.6 &98.2 &85.5 &66.2
\\
\mlp{}-3 &85.1 &97.9 &84.9 &53.8
\\
\sidetune{} &84.7 &96.9 &85.8 &48.6
\\
\bias{} &88.4 &98.8 &\bfseries91.2 &79.4
\\
\vptsh{} &86.7 &98.4 &90.7 &68.7
\\
\vpt{} &\bfseries88.5 &\bfseries99.0 &90.2 &83.6
\\
\expres{} (\textit{ours})&88.3&\bfseries99.0&90.0&80.5
\\
\bottomrule
\end{tabular}
\end{center}
\caption{\textbf{FGVC benchmark:} Per task adaptation results with ViT-B/16 model.}
    \label{tab:fgvc}
\end{table}
\noindent In Table \ref{tab:fgvc}, we compare the performance of our approach to adaptation baselines  and other prompting techniques on the FGVC benchmark. On four datasets our method outperforms most baselines and performs competitively with other prompting techniques.

\section{Additional Discussion on Performance-Cost Tradeoff}
\begin{figure}[h]
\centering
\includegraphics[width=0.9\columnwidth]{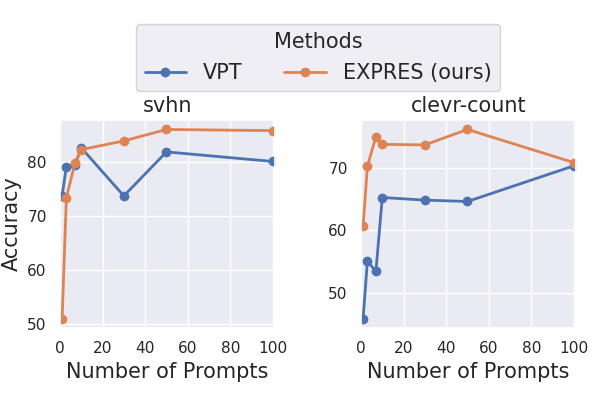}
    \caption{\textbf{Additional Computational Efficiency of EXPRES} Comparing Accuracy vs Number of Prompts for VPT and \expres{}}
    \label{fig:compef_supp}
\end{figure}

\noindent In Figure \ref{fig:compef_supp}, we compare the computational efficiency of our \expres{} with VPT on additional \vtab{} datasets. We fix the learning rate and weight decay for EXPRES at $lr=0.01, wd=0.0001$ and for VPT at the optimal values from \cite{jia2022vpt}. Rest all hyperparameters like batch size, epochs and warmup epochs are left unchanged. We observe trends similar to Figure \ref{fig:compef}, where for a given accuracy specification, the difference between number of prompts required for \expres{} and VPT can be upto an order (\eg, clevr-count in Figure \ref{fig:compef_supp}).

\begin{table}[t]
    \centering
    \footnotesize
    \begin{tabular}{c|cc}
         & Tuned Params ($\%$) & GMACs \\
         \midrule
         FT-all & 100.0 &17.47\\
         Linear & 0.090 &17.47\\
         \midrule
         \vptsh{} (M=1)& 0.091&17.56\\
         \vptsh{} (M=100)& 0.179& 26.87\\
         \midrule                  
         \vpt{} (M=1)& 0.100&17.50\\
         \vpt{} (M=100)& 1.166&19.96\\
         \midrule
         \expres{} (M=1)& 0.144 &17.56\\
         \expres{} (M=100)& 5.560 & 26.87\\
         \midrule
    \end{tabular}
    \caption{\textbf{Memory and computational cost analysis using a ViT-B/16 pre-trained on supervised ImageNet-21k}. We consider input resolution of $224\times224\times3$ and a $100$-way classification task. Tuned parameters are reported as a percentage of the parameters in the backbone model (including classifier head). Here, M is number of prompts per layer.}
    \label{tab:cost}
\end{table}

\noindent In Table \ref{tab:cost}, we compare the computational as well as memory cost of various adaptation techniques. The computational cost is reported in terms of GMACs and memory cost in terms of tuned parameters relative to full model parameters. We observe that with $100$ prompts, our \expres{} requires memory ($4.7$M params.) that is comparable to \vpt{} ($1$M params.) and  orders of magnitude less than finetuning  ($\sim 86$M params.). While the computational cost of our method ($26.9$ GMACs) is slightly more than Linear ($17.5$ GMACs) and \vpt{} ($19.9$ GMACs) when using large number ($100$) of prompts, our method greatly outperforms Linear (Table \ref{tab:vtab}) and achieves better optimal performance than \vpt{} with far fewer prompts (Figure \ref{fig:compef} and Figure \ref{fig:compef_supp}), providing good performance-cost tradeoffs.

\section{Ablations for Semantic Segmentation}
\begin{table}[h]
    \centering
    \scriptsize
    \begin{tabular}{cccc}
    \toprule
    &Q&K&MLP\\
    \midrule
    Accuracies &44.39&\bfseries51.61& 38.25\\
    \bottomrule
    \end{tabular}
\caption{\textbf{Ablation:} Effect of various representations for five-shot semantic segmentation on \pascal{0}}
\label{tab:segout}
\end{table}
\noindent We evaluate various ways of the extracting dense representations from transformer backbone for five-shot semantic segmentation. In Table \ref{tab:segout}, we compare last-layer keys ($K$), queries ($Q$) and MLP-block($MLP$) outputs. We use a learning rate of $0.005$, weight decay of $0.0001$, $5$ prompts and average the accuracies over $100$ tasks randomly sampled from the train-split of \pascal{0}.
We observe that keys (K) are the most effective representations for semantic segmentation, so we use them in all our segmentation experiments.

\section{Interpretability of Learnt  Prompts}
\begin{figure}[h]
       \includegraphics[width=1\columnwidth]{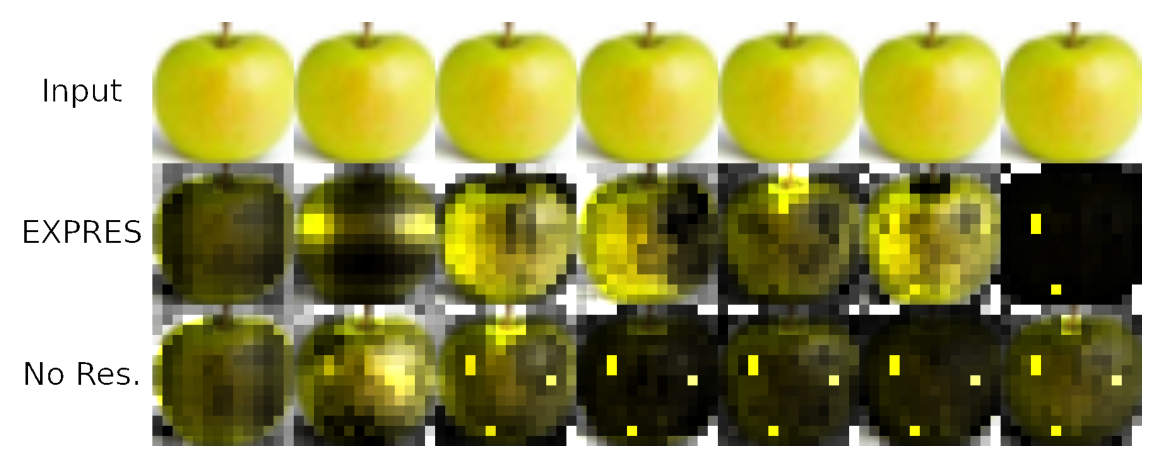}
        \caption{\textbf{\expres{} prompt attention at different layers:} We display input image (top-row), the attention maps of a trained \expres{} prompt (middle-row), and the attention map of the same prompt without residuals (bottom-row) evaluated at different layers. Attention maps for this prompt from early (close to input) to later (close to output) layers are arranged from left to right in each row.}
     \label{fig:modulate}
 \end{figure}
 
\noindent In Figure \ref{fig:modulate}, we provide visualizations to demonstrate that residual prompts learn semantic information and facilitate fine-grained layerwise modulation of the attention.
Here, an arbitrarily chosen but fixed prompt location is used for visualization purposes. We observe that residual prompts learn spatially fine-grained details that are diverse across layers and that removing them reduces the diversity of the attention maps across layers, confirming our hypothesis.


\end{document}